\documentclass{article}
\usepackage{times}

\usepackage{multirow}
\usepackage{multicol}
\usepackage[bookmarks=true]{hyperref}
\usepackage{booktabs}
\usepackage{array}
\usepackage{amssymb}
\usepackage{amsmath}
\usepackage{graphicx}
\usepackage{capt-of}
\usepackage{makecell}
\usepackage{xcolor}
\usepackage{wrapfig}
\usepackage{caption}
\usepackage{longtable}
\usepackage{tabularx}
\usepackage{adjustbox}

\usepackage{enumitem}

\newcolumntype{C}[1]{>{\centering\arraybackslash}p{#1}}

\usepackage[final]{corl_2026} 

\definecolor{citecolor}{HTML}{0071BC}
\hypersetup{colorlinks,linkcolor={red},citecolor={citecolor}}

\title{When Vision Overrides Language: Evaluating and Mitigating Counterfactual Failures in VLAs}

\author{
\adjustbox{max width=0.95\textwidth}{%
Yu Fang, Yuchun Feng, Dong Jing, Jiaqi Liu,
Yue Yang, Zhenyu Wei, Daniel Szafir, Mingyu Ding
}\\
University of North Carolina at Chapel Hill \\
{\href{https://vla-cf.github.io/}{\textcolor{blue}{https://vla-cf.github.io/}}}
}

\begin{document}
\maketitle


\begin{figure}[h]
  \centering
  \vspace{-8pt}
   \includegraphics[width=0.99\linewidth]{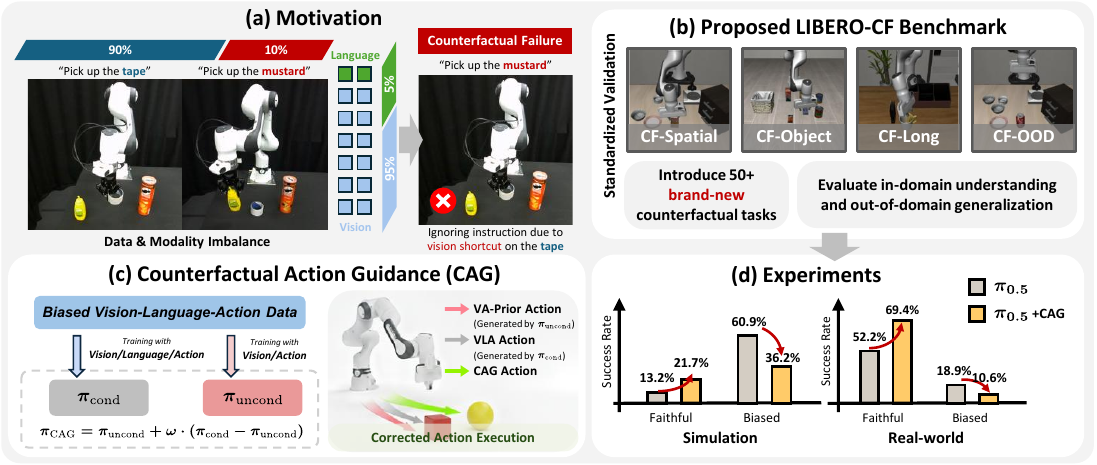}
   \caption{\textbf{Overview.} (a) VLAs often suffer from \emph{counterfactual failures} due to \emph{vision shortcuts}, defaulting to well-learned scene-specific behaviors instead of faithfully following instructions. (b) We study this issue and introduce LIBERO-CF, \emph{the first counterfactual benchmark} for evaluating language following in VLAs. (c) We propose \emph{Counterfactual Action Guidance (CAG)}, a dual-branch inference scheme that mitigates counterfactual failures in VLAs. (d) Experimental comparison.}
   \vspace{-8pt}
   \label{fig:teaser}
\end{figure}

\begin{abstract}
    Vision-Language-Action models (VLAs) promise to ground language instructions in robot control, yet in practice often fail to follow language faithfully.
    When presented with instructions that lack strong scene-specific supervision, VLAs suffer from \emph{counterfactual failures}:
    they act based on \emph{vision shortcuts} induced by dataset biases, repeatedly executing well-learned behaviors and selecting objects frequently seen during training regardless of language intent. 
    To support systematic study, we introduce LIBERO-CF, the first counterfactual benchmark for VLAs that evaluates language following capability by assigning alternative, feasible task instructions under the same LIBERO scene layouts.
    Our evaluation reveals that counterfactual failures are prevalent yet underexplored across state-of-the-art VLAs.
    To mitigate these failures, we propose Counterfactual Action Guidance (CAG), a simple yet effective dual-branch inference scheme that explicitly strengthens language conditioning in VLAs.
    CAG combines a standard VLA policy with a parallel language-unconditioned branch, amplifying the difference between their action predictions induced by language instructions.
    This design reduces reliance on vision shortcuts, improves robustness on counterfactual tasks, and requires neither additional demonstrations nor modifications to the base VLA architecture or its pretrained weights.
    Extensive experiments demonstrate consistent improvements across diverse VLAs.
    On LIBERO-CF, our training-free variant, TF-CAG, improves $\pi_{0.5}$ by 9.7\% in language grounding and 3.6\% in task success.
    When using a separate Vision-Action (VA) module, VA-CAG achieves larger gains of 15.5\% and 8.5\%, respectively.
    In real-world evaluations, VA-CAG improves language grounding by 9.4\% and task success by 17.2\% on average. 
\end{abstract}



\section{Introduction}
	
Vision-Language-Action models (VLAs) have achieved impressive performance in general robotic manipulation by leveraging pretrained Vision-Language Models (VLMs) to ground language in perception and action~\cite{team2024octo,brohan2022rt,brohan2023rt,kim24openvla,black2024pi_0}.
Despite recent advances, we identify a prevalent and underexplored failure mode in VLAs: poor language following behavior. 
In particular, when presented with instructions that have limited scene-specific supervision, VLAs often disregard the given instructions and default to executing well-learned tasks in the scene or selecting objects frequently seen during training, termed as \emph{counterfactual failures}.
Such behavior raises critical concerns about the reliability of VLAs as general-purpose robotic agents, as they cannot faithfully follow user intent even in familiar environments, posing significant risks to safety and usability for real-world deployment.

We attribute this limitation to the nature of robotic datasets and model modality imbalance, which are typically task-specific and visually dominant.
Under a fixed scene, demonstrations are often collected for only a small subset of tasks, encouraging VLAs to rely on \emph{vision shortcuts} rather than faithfully grounding language.
As a result, language conditioning may have insufficient influence on action selection, hindering VLAs from fully realizing the generalization capabilities inherited from pretrained VLMs.
Fig.~\ref{fig:vision_shortcut} provides direct evidence of this phenomenon: even under counterfactual or empty instructions, VLAs tend to execute the training task associated with the scene, and removing the training-task object significantly improves counterfactual performance.
Despite these observations, the field still lacks a systematic benchmark and a general solution for VLAs.

To systematically study language following behavior of VLAs, we introduce LIBERO-CF, the first counterfactual benchmark designed to evaluate counterfactual failures.
LIBERO-CF assigns alternative feasible language instructions under LIBERO scene layouts, covering different object instances, targeting objects that previously appeared only as background, long-horizon tasks, and out-of-distribution objects that are unseen during VLA finetuning.
Using this benchmark, we demonstrate that existing VLAs exhibit severe counterfactual failures: all the models perform poorly on newly assigned instructions while defaulting to executing well-learned training tasks.

To mitigate counterfactual failures, we propose Counterfactual Action Guidance (CAG), a dual-branch inference scheme that enhances language conditioning by mixing actions between a standard VLA and a language-unconditional vision-action (VA) policy.
CAG is plug-and-play, compatible with diverse VLA architectures, and requires no modifications to existing model architectures or pretrained weights.
Through extensive simulation and real-world experiments, we demonstrate consistent improvements over diverse state-of-the-art VLAs in both language grounding and task success across spatial, object, goal, long-horizon, and out-of-distribution instruction settings. 

To summarize, our contributions are threefold.
1) We introduce LIBERO-CF, the first counterfactual benchmark, designed to evaluate language-following in VLAs by distinguishing visually plausible yet linguistically distinct instructions under controlled counterfactual settings.
2) We propose Counterfactual Action Guidance, a dual-branch inference scheme that enhances language following for diverse VLAs, without changing existing architectures or pretrained models.
3) We provide a comprehensive analysis of vision shortcuts and counterfactual failures in VLAs, and demonstrate consistent improvements over state-of-the-arts through both LIBERO-CF and real-world experiments.


\section{Related Work}

\noindent \textbf{Vision-Language-Action Models.}
Vision-Language-Action (VLA) models have rapidly emerged as a promising framework for general-purpose robotic manipulation~\cite{brohan2022rt,brohan2023rt,Zhao2023ACT,li2023vision,wu2023unleashing,cheang2025gr,bjorck2025gr00t,liu2024rdt,chen2024moto,fan2025interleave,cui2025openhelix,cen2025worldvla,zheng2024tracevla,qu2025spatialvla,li2025spatial,liu2025hybridvla,deng2025graspvla,shi2025hi,zhong2025flowvla,dreamvla25,wang2025unified,cen2025rynnvla,tang2025incentivizing,jing2025mixture_of_horizons,li2025simplevla,zheng2025xvla,li2026watchact}.
They predict robot actions from visual observations and language instructions, enabling closed-loop control by repeatedly executing actions and updating observations.
Recent work has explored different model architectures and action representations, including VLM-based policies, world-model-based policies, discrete action tokens, continuous actions, flow matching, and diffusion-based action generation~\cite{paligemma,karamcheti2024prismatic,starvla2025,pi0-fast,chi2023diffusion,achiam2023gpt,zhu2023minigpt,yang2024qwen2,zhu2025internvl3,jing2024fineclip,liang2025video,jang2025dreamgen,shang2025roboscape}.
Despite these advances, our work shows that strong performance on standard benchmarks can still coexist with weak counterfactual language grounding, where policies rely on scene-specific visual cues rather than faithfully following the instruction.

\noindent \textbf{Modality Imbalance in MultiModal Models.}
Modality Imbalance has been widely studied in Visual Question Answering (VQA) and Multimodal Large Language Models (MLLMs)~\cite{niu2021counterfactual,udomcharoenchaikit2022mitigating,liu2024paying,schrodi2024two,zhang2025one,zhang2025cf,zheng2025mllms}.
Common mitigation strategies include counterfactual data augmentation~\cite{zhang2025cf}, modality reweighting~\cite{zheng2025mllms}, and training-time regularization~\cite{niu2021counterfactual} to encourage balanced multimodal reasoning.
Modality imbalance in VLAs remains underexplored, as they remain prone to losing instruction-following capabilities~\cite{kim2025openvla-oft}.
Recent works have explored introducing vision-language co-training~\cite{intelligence2025pi_}, instruction augmentation~\cite{fei2025libero,glossop2025cast,zhou2025libero}, architectural refinements~\cite{intelligence2025pi_}, instruction-following evaluation~\cite{fang2025intention}, and test-time verification~\cite{kwok2026scaling}.
However, the field still lacks systematic study of language following and a universal solution to mitigate counterfactual failures.
We address both gaps in this paper.
	

\section{Problem Formulation}
\subsection{Vision Bias in Datasets and Architectures}
Large-scale robotic datasets have driven recent advances in VLAs~\cite{o2023open,khazatsky2024droid,bu2025agibot}.
However, textual signal is often less diverse than visual and action, and limited vision-language supervision may weaken the language capabilities inherited from pretrained VLMs~\cite{intelligence2025pi_}.
This imbalance can encourage vision shortcuts, where VLAs rely on scene-specific visual cues rather than grounding language instructions~\cite{fei2025libero,glossop2025cast,zhou2025libero,xu2025seeing,peng2025counterfactual,zhan2026stable,lian2026bayesianvla}.
On the other hand, in typical VLAs, language tokens (typically dozens) are vastly outnumbered by visual tokens (often hundreds/thousands), creating a risk where textual influence can be marginalized.
While most VLAs are built upon pre-trained VLMs with strong vision-language priors, the alignment among the three modalities may still be insufficient.

\subsection{Problem Formulation from Bayesian Perspective}
\label{sec:formulation}
We provide a conceptual view of how visual observations and language instructions jointly influence action prediction in VLA policies.
Given an observation $o$ and a language instruction $l$, the action distribution can be viewed through a Bayesian perspective:
\begin{equation}
P(a \mid o, l) \propto P(a \mid o) \cdot P(l \mid a, o),
\label{eq:bayes}
\end{equation}
where $P(a \mid o)$ denotes the vision-conditioned action prior, and $P(l \mid a,o)$ captures how compatible an action is with the observation and language instruction.

In principle, language conditioning should shift the vision-conditioned action prior toward actions consistent with the instruction.
However, in scene-specific manipulation, visual observation alone can often provide strong cues about the expected task.
In Tab.~\ref{tab:libero}, the evaluated VLAs retain substantial task performance when language inputs are removed.
This suggests that VLAs may rely heavily on the vision-conditioned action prior, with language providing insufficient influence to override scene-specific visual biases.
We refer to this failure mode as \emph{vision shortcut}.

We attribute this behavior to a common modality imbalance during VLA training.
In addition to vision bias in robotic datasets, consider a typical scenario that, under a fixed visual observation $o$, there exists a set of feasible tasks $L_o=\{l_o^1, l_o^2, \cdots, l_o^n\}$. 
During VLA finetuning for a specific scene, only a limited subset $L_o^{\text{in}} \subset L_o$ is collected with sufficient expert demonstrations.
The remaining tasks $L_o^{\text{out}} = L_o \setminus L_o^{\text{in}}$ are under-observed during this scene-specific adaptation, even if they appear in large-scale pretraining.
As a result, the observation $o$ becomes strongly associated with tasks in $L_o^{\text{in}}$, potentially biasing action predictions away from instructions in $L_o^{\text{out}}$, leading to vision shortcuts.

In this paper, we investigate predicted actions $a$ of a trained VLA policy $\pi(a|o,l)$, under different language instructions $l \in L_o$ and the same visual observation $o$.
Notably, our problem formulation differs from prior settings that rephrase language instructions within the same task~\cite{fei2025libero}.
Instead, we study counterfactual instructions and directly probe language following behaviors in VLAs.

\begin{figure*}[t]
  \centering
   \includegraphics[width=0.99\linewidth]{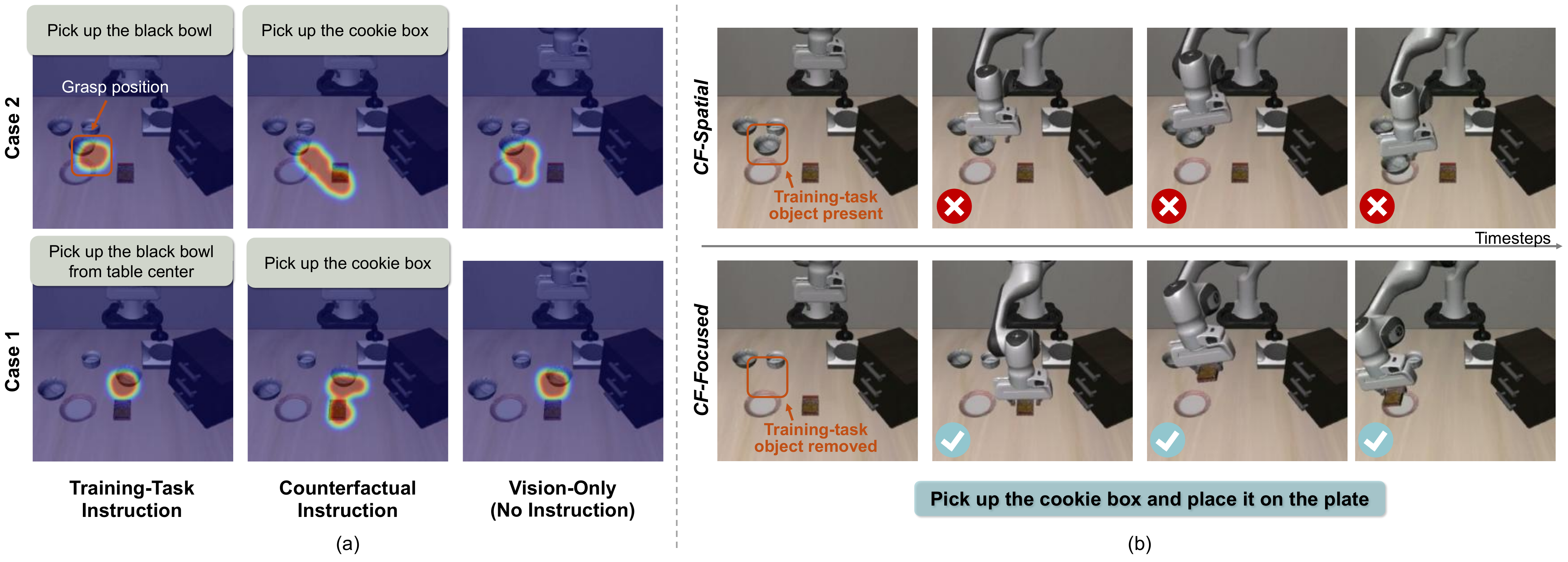}
   \vspace{-4pt}
   \caption{\textbf{Evidence for Vision Shortcuts in VLAs.} (a) We visualize the distribution of grasp positions from 50 trials as heatmaps under different instructions.
   Even when given counterfactual or empty instructions, VLAs execute the well-learned training task in the scene.
   (b) Removing the training-task object in the scene improves the success rates of VLAs on counterfactual instructions.}
   \vspace{-12pt}
   \label{fig:vision_shortcut}
\end{figure*}

\section{The LIBERO-CF Benchmark}
We present LIBERO-CF, a counterfactual benchmark built on LIBERO scene layouts~\cite{liu2023libero}.
LIBERO is a widely used robotic manipulation benchmark for VLA finetuning and evaluation, but its tasks are in-domain with respect to the training demonstrations.
In contrast, LIBERO-CF evaluates whether finetuned VLAs can follow \emph{counterfactual task instructions}: alternative feasible tasks under the same layouts that are under-observed or entirely unseen during finetuning.

\textbf{Suites.}
We construct four suites:
1) \emph{CF-Spatial} (15 tasks): Evaluates spatial language grounding by targeting original background objects.
2) \emph{CF-Object} (10 tasks): Evaluates alternative instructions targeting different objects.
3) \emph{CF-Long} (10 tasks): Evaluates long-horizon language following involving new targets.
4) \emph{CF-OOD} (15 tasks): Evaluates generalization with unseen objects.

\textbf{Metrics.}
To quantitatively evaluate language following behaviors, we define a \emph{grounding rate} (``grounding"), which measures whether the gripper makes contact with the target object specified in the instruction, regardless of task completion.
This metric reflects how faithfully the model follows the given language instructions.
In addition, we report the \emph{success rate} (``success"), which measures whether the robot completes the task, providing a stricter measure of task execution.

\textbf{Evaluation Dimensions.}
In addition to reporting performance, we quantify how often a VLA defaults to well-learned scene-specific behaviors. 
In the tables, we denote ``Faithful" when the model follows the instructed task, and ``Biased" when it instead follows the original training task.

\subsection{Validating Vision Shortcuts of VLAs}
\textbf{Baselines.} 
We discuss language-following ability of representative VLAs, including OpenVLA-OFT~\cite{kim2025openvla-oft}, 
$\pi_0$~\cite{black2024pi_0}, and  
$\pi_{0.5}$~\cite{intelligence2025pi_}.
We use their official models on LIBERO in our evaluations.

\begin{wraptable}{r}{0.45\linewidth}
\vspace{-15pt}
\centering
\scriptsize
\setlength{\tabcolsep}{2pt}
\renewcommand{\arraystretch}{0.85}
\caption{\textbf{Evidence of vision shortcuts on LIBERO.} Results are success rates (\%).}
\label{tab:libero}
\begin{tabular}{
  >{\centering\arraybackslash}p{0.9cm}
  >{\centering\arraybackslash}p{0.75cm}
  >{\centering\arraybackslash}p{0.65cm}
  >{\centering\arraybackslash}p{0.65cm}
  >{\centering\arraybackslash}p{0.65cm}
  >{\centering\arraybackslash}p{0.65cm}
  >{\centering\arraybackslash}p{0.65cm}
}
\toprule
\textbf{Model} & \textbf{Input} & \textbf{Spatial} & \textbf{Object} & \textbf{Goal} & \textbf{Long} & \textbf{Avg.}\\
\midrule
\multirow{2}{*}{\makecell{OpenVLA\\-OFT}}
  & V+L    & 97.6 & 98.4 & 97.9 & 94.5 & 97.1 \\
  & V-only & 83.2 & 98.2 & 10.0 & 85.0 & 69.1 \\
\midrule
\multirow{2}{*}{$\pi_0$}
  & V+L    & 96.8 & 98.8 & 95.8 & 85.2 & 94.2 \\
  & V-only & 28.0 & 26.8 & 3.4  & 15.6 & 18.5 \\
\midrule
\multirow{2}{*}{$\pi_{0.5}$}
  & V+L    & 98.8 & 98.2 & 98.0 & 92.4 & 96.9 \\
  & V-only & 68.2 & 56.6 & 10.2 & 71.6 & 51.7 \\
\bottomrule
\end{tabular}
\vspace{-10pt}
\end{wraptable}

\textbf{Quantitative Results.}
We validate vision shortcuts by comparing standard vision-language inputs with vision-only inputs on LIBERO.
In Tab.~\ref{tab:libero}, all evaluated VLAs retain high performance after removing language, indicating that vision alone can induce strong task-specific action preferences.
This is especially pronounced for OpenVLA-OFT and $\pi_{0.5}$, which achieve 69.1\% and 51.7\% average success under vision-only inputs.
LIBERO-Goal shows weaker vision-only performance because the scene is fixed, and task goals vary primarily through language.

\textbf{Grasping Distributions.}
In Fig.~\ref{fig:vision_shortcut}(a), we visualize the grasp positions from 50 trials on pick-and-place tasks in \emph{CF-Spatial} under different language instructions using $\pi_{0.5}$ at inference time.
When given the training-task instruction, grasp positions are tightly clustered around the corresponding training-task object, suggesting the task is well-learned.
In contrast, under counterfactual instructions, the grasping distribution spreads not only around the instructed object but also around the training-task object.
When given no instruction, the distribution still primarily centers on the training-task object.
These results provide further evidence that VLAs tend to default to executing well-learned training tasks even when presented with alternative language instructions, offering insight into the nature of vision shortcuts and counterfactual failures.

\begin{wraptable}{r}{0.48\linewidth}
\vspace{-10pt}
\centering
\scriptsize
\setlength{\tabcolsep}{2pt}
\renewcommand{\arraystretch}{0.95}
\caption{\textbf{Impact of training-task objects on vision shortcuts.} Results are percentages (\%).}
\label{tab:training_task_objects}
\begin{tabular}{
    >{\centering\arraybackslash}p{1.4cm}
    >{\centering\arraybackslash}p{1.4cm}
    >{\centering\arraybackslash}p{1.4cm}
    >{\centering\arraybackslash}p{1.4cm}
}
\toprule
\textbf{Model} & \textbf{Metric} & \textbf{CF-Spatial} & \textbf{CF-Focused} \\
\midrule
\multirow{2}{*}{\makecell{OpenVLA\\-OFT}}
    & Grounding & 6.8  & \textbf{31.9} \\
    & Success & 1.1  & \textbf{14.4} \\
\midrule
\multirow{2}{*}{$\pi_0$}
    & Grounding & 70.9 & \textbf{92.7} \\
    & Success & 34.4 & \textbf{51.1} \\
\midrule
\multirow{2}{*}{$\pi_{0.5}$}
    & Grounding & 39.3 & \textbf{68.5} \\
    & Success & 24.4 & \textbf{54.4} \\
\hline
\addlinespace[1pt]
\hline
\noalign{\vskip 3pt}
\multirow{2}{*}{Avg.}
    & Grounding & 39.0 & \textbf{64.4} \\
    & Success & 20.0 & \textbf{40.0} \\
\bottomrule
\end{tabular}
\vspace{-10pt}
\end{wraptable}

\textbf{Impact of Training-task Objects.}
We validate that counterfactual failures are related to visual distraction.
We construct a new evaluation suite termed \emph{CF-Focused},  by removing training-task objects from the scenes in \emph{CF-Spatial}.
As shown in Fig.~\ref{fig:vision_shortcut}(b) and Tab.~\ref{tab:training_task_objects}, we compare VLA performance on the two suites.
Although both suites use identical counterfactual instructions, all VLAs show improved performance on \emph{CF-Focused}.
On average, this improves language grounding from 39.0\% to 64.4\% and doubles task success
from 20.0\% to 40.0\%.
These results further validate the presence of vision shortcuts in VLAs and indicate that training-task objects act as strong visual attractors, biasing policies toward
well-learned behaviors despite counterfactual instructions.

\textbf{Summary.}
These observations provide evidence of \emph{vision shortcuts} in VLAs, and motivate the need for counterfactual evaluation and guidance to strengthen language conditioning.

\section{Method}

\subsection{Counterfactual Action Guidance}
\begin{wrapfigure}{r}{0.52\textwidth}
    \vspace{-14pt}
    \centering
    \includegraphics[width=\linewidth]{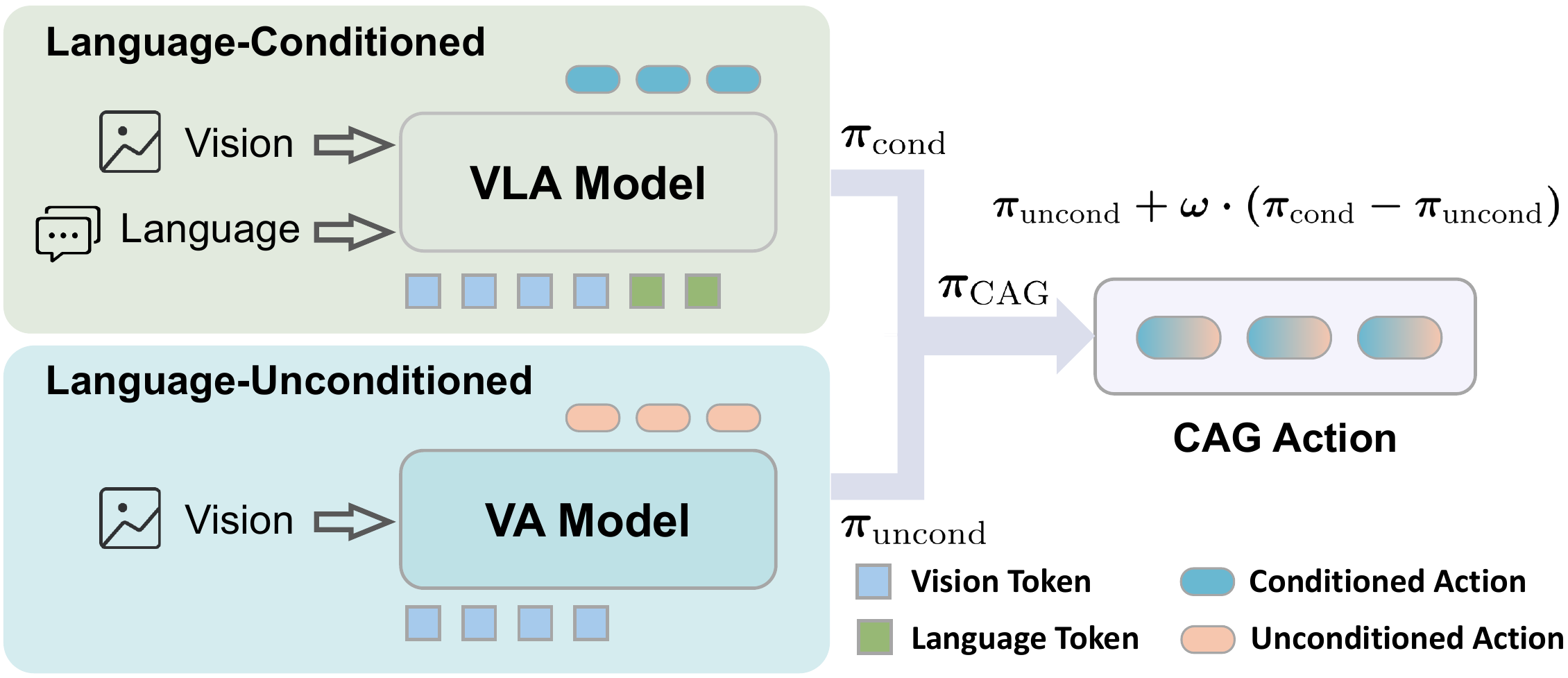}
    \vspace{-12pt}
    \caption{
        \textbf{CAG Overview.}
    }
    \label{fig:method}
    \vspace{-12pt}
\end{wrapfigure}
\label{sec:cag}
Inspired by classifier-free guidance~\cite{ho2022classifier}, we introduce counterfactual action guidance (CAG), an inference-time action mixing scheme for enhancing language conditioning in VLAs.
As illustrated by Eq.~\ref{eq:bayes}, language conditioning should shift vision-only action preferences toward the instructed behavior.
However, strong visual priors may dominate action predictions, leading to counterfactual failures.
To mitigate this issue, CAG estimates and amplifies the language-induced action shift beyond vision-only preferences.

Given an observation $o$ and instruction $l$, we consider two policies: a language-conditioned policy $\pi_{\text{cond}}(a \mid o, l)$ and a language-unconditioned (vision-only) policy $\pi_{\text{uncond}}(a \mid o)$, which produce predicted action chunks $a_{\mathrm{cond}}$ and $a_{\mathrm{uncond}}$, respectively.
Their difference 
$\Delta a_l =
a_{\mathrm{cond}} -
a_{\mathrm{uncond}}$
estimates the action shift associated with language
conditioning under the same visual observation.
We define the CAG action prediction as
\begin{equation}
a_{\mathrm{CAG}}
=
a_{\mathrm{uncond}}
+
\omega \left(
a_{\mathrm{cond}}
-
a_{\mathrm{uncond}}
\right),
\label{eq:cag}
\end{equation}
where $\omega \ge 0$ is a guidance scale controlling the strength of language conditioning.
By scaling this shift with $\omega$, CAG encourages the policy to follow the instruction when it conflicts with the behavior suggested by visual priors.
To ensure broad applicability across different VLA architectures, CAG operates directly on the final predicted action chunks after both branches complete their inference.
While Eq.~\ref{eq:bayes} provides the probabilistic motivation, CAG adopts a CFG-inspired extrapolation heuristic in action space.
This provides a unified inference-time action-space guidance mechanism for both flow-based and deterministic regression policies, without modifying the original VLA architectures or pretrained weights.

\subsection{Dual-branch Inference with CAG}
To enhance language grounding in VLAs as described in Eq.~\ref{eq:cag}, CAG requires a language-conditioned policy $\pi_{\text{cond}}$ and an unconditioned vision-only policy $\pi_{\text{uncond}}$ to estimate and amplify the action shift induced by the language instruction.
As shown in Fig.~\ref{fig:method}, we implement a CAG policy as a dual-branch inference scheme that combines a language-conditioned VLA with a vision-only branch.
We consider two strategies for obtaining the language-unconditioned branch:

\textbf{1) Training-Free Strategy (\emph{TF}).} 
Without additional finetuning efforts, we first consider directly using a standard VLA to approximate both $\pi_\text{cond}$ and $\pi_\text{uncond}$. 
Specifically, we obtain $\pi(a \mid o, \ell)$ with language input, and derive an approximate $\pi(a \mid o)$ by dropping language input at inference time.

\textbf{2) Training a Vision-Action Prior (\emph{VA}).}
We opt to train a separate Vision-Action (VA) Model to explicitly obtain $\pi(a \mid o)$.
Compared to standard VLA, VA provides a cleaner vision-only prior, which is decoupled from language supervision.
At inference time, VA is paired with the standard VLA within the same dual-branch guidance framework, resulting in improved language grounding and reduced vision shortcuts.

\begin{table*}[t]
    \scriptsize
    \caption{\textbf{Quantitative evaluation of VLAs on the LIBERO-CF benchmark.} Results are language grounding and task success rates (\%). F. denotes Faithful and B. denotes Biased.}
    \centering
    \renewcommand{\arraystretch}{0.85}
    \setlength{\tabcolsep}{3.2pt}
    \begin{tabular}{
      >{\centering\arraybackslash}p{1.8cm} |
      >{\centering\arraybackslash}p{1.1cm} |
      >{\centering\arraybackslash}p{0.7cm} |
      >{\centering\arraybackslash}p{0.7cm}
      >{\centering\arraybackslash}p{0.7cm}
      >{\centering\arraybackslash}p{0.7cm}
      >{\centering\arraybackslash}p{0.7cm}
      >{\centering\arraybackslash}p{0.7cm}
      >{\centering\arraybackslash}p{0.7cm}
      >{\centering\arraybackslash}p{0.7cm}
      >{\centering\arraybackslash}p{0.7cm}
      >{\centering\arraybackslash}p{0.7cm}
      >{\centering\arraybackslash}p{0.7cm}
    }
    \toprule
    \multirow{3}{*}{\textbf{Model}} &
    \multirow{3}{*}{\textbf{Metric}} &
    \multirow{3}{*}{\textbf{CAG}}
    & \multicolumn{2}{c}{\textbf{CF-Spatial}}
    & \multicolumn{2}{c}{\textbf{CF-Object}}
    & \multicolumn{2}{c}{\textbf{CF-Long}}
    & \multicolumn{2}{c}{\textbf{CF-OOD}}
    & \multicolumn{2}{c}{\textbf{Average}} \\
    \cmidrule(lr){4-5}\cmidrule(lr){6-7}\cmidrule(lr){8-9}
    \cmidrule(lr){10-11}\cmidrule(lr){12-13}
    & & & \textbf{F. $\uparrow$} & \textbf{B. $\downarrow$}
          & \textbf{F. $\uparrow$} & \textbf{B. $\downarrow$}
          & \textbf{F. $\uparrow$} & \textbf{B. $\downarrow$}
          & \textbf{F. $\uparrow$} & \textbf{B. $\downarrow$}
          & \textbf{F. $\uparrow$} & \textbf{B. $\downarrow$} \\
    \midrule

\multirow{6}{*}{\makecell{OpenVLA-OFT\\~\cite{kim2025openvla-oft}}}
& \multirow{3}{*}{Grounding}
& /        & 6.8  & 84.7 & 2.0 & 99.6 & 4.0 & 62.9 & 6.0 & 87.3 & 4.7 & 83.6 \\
& & TF & 11.1 & 78.5 & 1.6 & 100.0 & 4.9 & 61.1 & 5.6 & 88.1 & 5.8 & 81.9 \\
& & VA & \textbf{26.5} & \textbf{59.5} & \textbf{3.8} & \textbf{91.8} & \textbf{6.9} & \textbf{60.0} & \textbf{7.9} & \textbf{50.1} & \textbf{11.3} & \textbf{65.4} \\
\cmidrule(lr){2-13}
& \multirow{3}{*}{Success}
& /        & 1.1 & 80.7 & 0.0 & 97.6 & 0.2 & 54.4 & 0.1 & 81.7 & 0.4 & 78.6 \\
& & TF & 3.3 & 73.3 & 0.0 & 94.4 & 0.2 & 54.0 & 0.1 & 83.5 & 0.9 & 76.3 \\
& & VA & \textbf{7.9} & \textbf{35.2} & 0.0 & \textbf{57.0} & \textbf{0.4} & \textbf{48.9} & 0.1 & \textbf{22.7} & \textbf{2.1} & \textbf{41.0} \\
\midrule

\multirow{6}{*}{$\pi_0$~\cite{black2024pi_0}}
& \multirow{3}{*}{Grounding}
& /        & 70.9 & 37.1 & 6.0 & 59.4 & 20.4 & 83.1 & 18.0 & 72.4 & 28.8 & 63.0 \\
& & TF & 74.5 & 32.5 & 12.2 & \textbf{50.6} & 19.6 & \textbf{81.8} & 21.1 & 72.4 & 31.9 & 59.3 \\
& & VA & \textbf{85.2} & \textbf{24.4} & \textbf{13.8} & 53.8 & \textbf{22.7} & 82.2 & \textbf{27.6} & \textbf{47.5} & \textbf{37.3} & \textbf{52.0} \\
\cmidrule(lr){2-13}
& \multirow{3}{*}{Success}
& /        & 34.4 & 19.1 & 0.0 & 45.0 & 1.1 & 63.1 & \textbf{2.7} & 52.9 & 9.6 & 45.0 \\
& & TF & 33.5 & 14.4 & \textbf{0.6} & \textbf{15.0} & 2.0 & \textbf{38.7} & 2.1 & 42.3 & 9.6 & 27.6 \\
& & VA & \textbf{36.7} & \textbf{6.1} & 0.0 & 29.2 & \textbf{2.9} & 48.7 & 2.1 & \textbf{19.2} & \textbf{10.4} & \textbf{25.8} \\
\midrule

\multirow{6}{*}{$\pi_{0.5}$~\cite{intelligence2025pi_}}
& \multirow{3}{*}{Grounding}
& /        & 39.3 & 61.3 & 11.4 & 68.8 & 51.6 & 58.4 & 20.7 & 74.0 & 30.8 & 65.6 \\
& & TF & \textbf{52.0} & 45.3 & 24.6 & 65.2 & 60.7 & \textbf{56.7} & 24.8 & 71.9 & 40.5 & 59.8 \\
& & VA & 50.7 & \textbf{45.2} & \textbf{34.0} & \textbf{49.8} & \textbf{64.2} & 61.1 & \textbf{36.4} & \textbf{52.8} & \textbf{46.3} & \textbf{52.2} \\
\cmidrule(lr){2-13}
& \multirow{3}{*}{Success}
& /        & 24.4 & 56.9 & 5.8 & 66.8 & 15.8 & 50.4 & 6.9 & 69.6 & 13.2 & 60.9 \\
& & TF & 27.3 & \textbf{33.3} & 9.2 & 41.4 & 20.9 & \textbf{36.7} & 9.9 & 62.7 & 16.8 & 43.5 \\
& & VA & \textbf{31.6} & 36.0 & \textbf{18.0} & \textbf{29.8} & \textbf{26.7} & 41.8 & \textbf{10.3} & \textbf{37.1} & \textbf{21.7} & \textbf{36.2} \\

\bottomrule
\end{tabular}
\vspace{-12pt}
\label{tab:libero_cf}
\end{table*}

\section{Simulation Experiments}
\label{sec:sim}

\subsection{Evaluating Counterfactual Behaviors}
\label{sec:counterfactual_behaviors}

\textbf{Performance of Baselines.}
In Tab.~\ref{tab:libero_cf}, we observe that all evaluated VLAs suffer from counterfactual failures.
OpenVLA-OFT exhibits the most severe vision shortcuts, achieving only a 4.7\% grounding rate and 0.4\% success rate.
Even $\pi_{0.5}$, which reaches the best overall performance among the baselines, achieves only a 30.8\% grounding rate and 13.2\% success rate.
In contrast, they maintain high performance on biased metrics: even with counterfactual instructions, OpenVLA-OFT achieves 83.6\% grounding and 78.6\% success on original training tasks, while $\pi_{0.5}$ achieves 65.6\% grounding and 60.9\% success.
This highlights the severity of counterfactual failures in current VLAs.
Implementation details and additional results on X-VLA~\cite{zheng2025xvla} are provided in Appendix~\ref{sec:supl_sim}.

\textbf{Effectiveness of CAG.}
We validate CAG with both variants: \emph{TF} and \emph{VA}.
In Tab.~\ref{tab:libero_cf}, compared to VLA baselines, both CAG variants consistently improve grounding and success rates on counterfactual tasks while decreasing biased execution of the original training tasks.
This demonstrates that CAG effectively mitigates vision shortcuts and corrects counterfactual failures by amplifying the language-induced action shift.
For example, on $\pi_{0.5}$, TF improves the average grounding rate from 30.8\% to 40.5\%, while VA further boosts it to 46.3\% with success rates increasing from 13.2\% to 21.7\%.
Importantly, these improvements are accompanied by a substantial reduction in counterfactual failures: for $\pi_{0.5}$, VA reduces biased grounding by 13.4\% and biased success by 24.7\%.
Notably, $\pi_{0.5}$ with VA even achieves 36.4\% grounding and 10.3\% success on CF-OOD, demonstrating that vision shortcuts significantly suppress the generalization capabilities of VLAs.
While our training-free strategy is already proven effective, VA generally yields stronger improvements, suggesting that explicitly modeling the vision-only prior provides a more reliable counterfactual correction than reusing a language-conditioned VLA policy as the unconditioned branch.

\subsection{Ablation Studies}

\begin{wraptable}{r}{0.5\textwidth}
    \vspace{-12pt}
    \centering
    \scriptsize
    \caption{\textbf{Training strategies for Counterfactual Action Guidance with $\pi_{0.5}$.}}
    \label{tab:training_strategies}
    \setlength{\tabcolsep}{2.5pt}
    \renewcommand{\arraystretch}{0.85}
    \resizebox{\linewidth}{!}{
    \begin{tabular}{
        >{\centering\arraybackslash}p{1.2cm}
        >{\centering\arraybackslash}p{0.8cm}
        >{\centering\arraybackslash}p{0.68cm}
        >{\centering\arraybackslash}p{0.68cm}
        >{\centering\arraybackslash}p{0.68cm}
        >{\centering\arraybackslash}p{0.68cm}
    }
    \toprule
    \multirow{2}{*}{\textbf{Metric}}
    & \multirow{2}{*}{\textbf{Strategy }}
    & \multicolumn{2}{c}{\textbf{CF-Spatial}}
    & \multicolumn{2}{c}{\textbf{CF-Long}} \\
    \cmidrule(lr){3-4} \cmidrule(lr){5-6}
    & & \textbf{F.} & \textbf{B.}
      & \textbf{F.} & \textbf{B.} \\
    \midrule
    
    \multirow{5}{*}{Grounding}
        & R1 & 39.3 & 61.3 & 51.6 & 58.4 \\
        & R2 & 9.5  & 89.3 & 34.0 & 83.1 \\
        & R3 & 52.0 & 45.3 & 60.7 & 56.7 \\
        & R4 & \textbf{52.3} & 47.5 & \textbf{64.9} & 59.6 \\
        & R5 & 50.7 & 45.2 & 64.2 & 61.1 \\
    \midrule
    
    \multirow{5}{*}{Success}
        & R1 & 24.4 & 56.9 & 15.8 & 50.4 \\
        & R2 & 3.7  & 86.5 & 2.7  & 71.3 \\
        & R3 & 27.3 & 33.3 & 20.9 & 36.7 \\
        & R4 & 27.7 & 37.1 & 26.4 & 41.6 \\
        & R5 & \textbf{31.6} & 36.0 & \textbf{26.7} & 41.8 \\
    \bottomrule
    \end{tabular}
    }
    \vspace{-12pt}
\end{wraptable}

\textbf{Training Strategies.}
We conduct ablation studies to investigate effective training strategies for CAG, using $\pi_{0.5}$ on CF-Spatial and CF-Long.
Using the vanilla $\pi_{0.5}$ as the baseline (\emph{R1}), we evaluate and compare several CFG-based variants:
\emph{R2}: a VLA finetuned with random language dropout as both $\pi_{\text{cond}}$ and $\pi_{\text{uncond}}$;
\emph{R3}: a standard VLA used as both $\pi_{\text{cond}}$ and $\pi_{\text{uncond}}$ at inference time (aligned with \emph{TF});
\emph{R4}: a standard VLA as $\pi_{\text{cond}}$ paired with a LoRA-finetuned VA model as $\pi_{\text{uncond}}$;
\emph{R5}: a standard VLA as $\pi_{\text{cond}}$ paired with a fully finetuned VA model as $\pi_{\text{uncond}}$ (aligned with \emph{VA}).

In Tab.~\ref{tab:training_strategies}, even training-free strategy (\emph{R3}) improves both grounding and success rates, showing the effectiveness of CAG.
When introducing a VA, both R4 and R5 achieve stronger performance than R3, indicating that explicitly modeling the vision-only prior is more effective.
R4 and R5 achieve comparable performance, suggesting both LoRA and full finetuning are effective for training a VA.
R5 further improves success rates, e.g., by 4.3\% on CF-Spatial, achieving better manipulation accuracy.
Notably, while dropout-based training (R2) is common for CFG in generative models, it performs poorly on VLAs: language dropout weakens its conditioning, leading to severe degradation toward vision-dominated behaviors.

\begin{wrapfigure}{r}{0.47\textwidth}
    \vspace{-16pt}
    \centering
    \includegraphics[width=\linewidth]{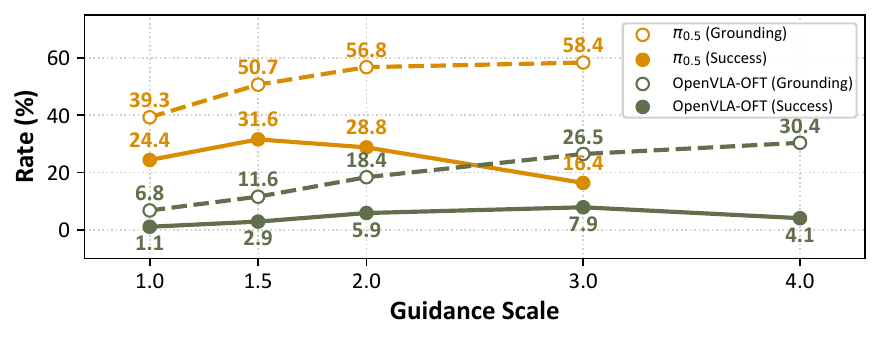}
    \vspace{-16pt}
    \caption{\textbf{Investigation of guidance scale.} }
    \label{fig:guidance_scale}
\end{wrapfigure}

\textbf{Guidance Scales.}
In Fig.~\ref{fig:guidance_scale}, we study the effect of guidance scale $\omega$ on $\pi_{0.5}$ and OpenVLA-OFT.
Increasing the guidance scale progressively mitigates vision shortcuts, while excessive guidance degrades manipulation accuracy.
OpenVLA-OFT benefits from a larger guidance scale, consistent with its stronger reliance on visual cues (see vision-only performance in Tab.~\ref{tab:libero}).
Guided by these observations, we empirically select a fixed guidance scale for each backbone to achieve the best manipulation performance: $\omega=1.5$ for $\pi_{0}$ and $\pi_{0.5}$, and $\omega=3.0$ for OpenVLA-OFT.

\section{Real-world Experiments}

\begin{figure*}[t]
  \centering
  \includegraphics[width=0.99\linewidth]{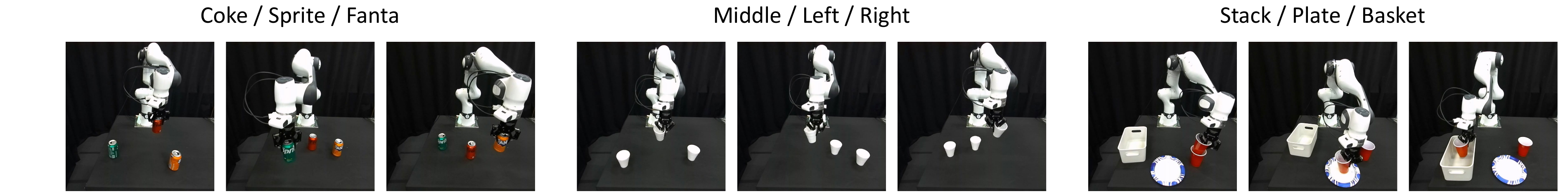}
  \vspace{2pt}
  \includegraphics[width=0.99\linewidth]{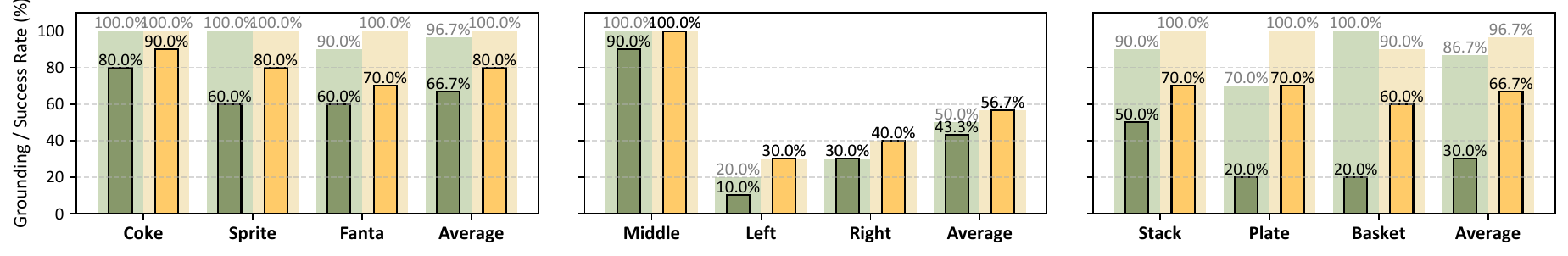}
  \includegraphics[width=0.99\linewidth]{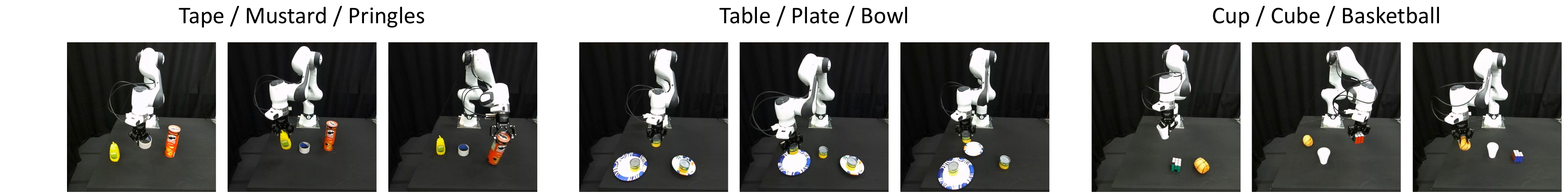}
  \includegraphics[width=0.99\linewidth]{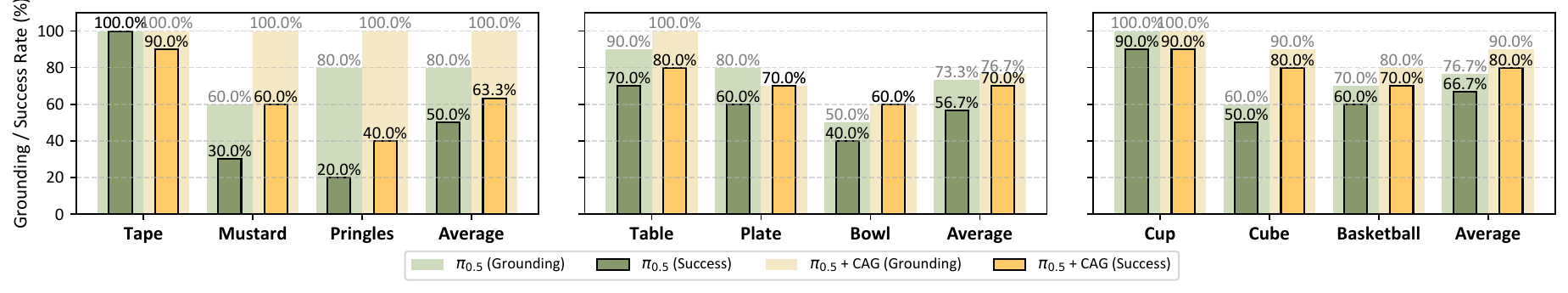}
  \vspace{-6pt}
  \caption{\textbf{Real-world experiments.} We study multiple aspects of language grounding, CAG consistently reduces counterfactual failures and improves task success across all scenes.}
  \label{fig:real-world-object-spatial-goal}
  \vspace{-16pt}
\end{figure*}

\subsection{Implementation Details}

\textbf{Task Design.} We study different aspects of language grounding in real-world settings, including \emph{Object Recognition}, \emph{Spatial Reasoning}, \emph{Goal Targeting}, and \emph{Out-of-Distribution Generalization (OOD)}.
Each scene is designed with three possible tasks: one well-learned in-domain task $L_o^{\text{in}}$ with sufficient expert demonstrations, two under-observed tasks $L_o^{\text{out}}$ with counterfactual instructions.

\textbf{Setup.} 
We use a Franka Research 3 with a Robotiq 2F-85 gripper.
Our setup follows the DROID platform~\cite{khazatsky2024droid}, including a ZED 2i as the exterior camera and a ZED mini as the wrist camera.
We randomly change the object positions and perform 10 trials per instruction for evaluation.

\textbf{Dataset and VLA Finetuning.}
To reflect the practical bias toward a dominant task, we collect 20 demonstrations for each $l \in L_o^{\text{in}}$.
For each $l \in L_o^{\text{out}}$, we collect only one demonstration for minimal warm-up, while setting OOD tasks in a zero-shot manner.
We set $\pi_{0.5}$ as our baseline and demonstrate the effectiveness of CAG with a VA model (aligned with \emph{VA} in Sec.~\ref{sec:counterfactual_behaviors}).
For each VLA and VA, we fully finetune the $\pi_{0.5}$-DROID model for 10k steps with a batch size of 32.

\subsection{Results}
In Fig.~\ref{fig:real-world-object-spatial-goal}, we present the results of our real-world evaluations.
For each scene, the first task is well-learned in the domain during finetuning, while the remaining two tasks are under-observed.

\textbf{Object Recognition.}
\emph{Coke/Sprite/Fanta} share identical geometry and differ only in visual appearance.
This represents a minimal counterfactual object recognition challenge that requires identical manipulation skills.
However, $\pi_{0.5}$ still exhibits grounding errors for under-observed tasks, indicating that vision shortcuts can hinder simple object recognition.
\emph{Tape/Mustard/Pringles} involves diverse object geometries and manipulation requirements.
$\pi_{0.5}$ performance drops sharply on two under-observed tasks, reflecting significant counterfactual failures.
CAG improves grounding and success rates on all under-observed tasks, while in-domain performance is preserved.
Notably, CAG achieves a 100\% average grounding rate, and improves the success rates by 13.3\% in both scenes.
These results highlight the existence of vision shortcuts and the effectiveness of CAG.

\textbf{Spatial Reasoning.}
In \emph{Middle/Left/Right} and \emph{Table/Plate/Bowl}, we respectively place three identical white cups and corn cans at different positions.
$\pi_{0.5}$ achieves only 20\% and 30\% grounding rates on under-observed targets (i.e., left, right) in \emph{Middle/Left/Right}, and 60\% for Plate and 40\% for Bowl in \emph{Table/Plate/Bowl}.
These show that VLAs are prone to vision shortcuts in spatially differentiated tasks.
CAG mitigates this failure mode, improving grounding by 16.6\% and task success by 13.3\%.

\textbf{Goal Targeting.}
In \emph{Stack/Plate/Basket}, we assign different goals to the same cup.
While $\pi_{0.5}$ shows decent grounding on under-observed goals, task success rates remain low.
It sometimes places the cup at an incorrect goal even when given training-task instruction, evidencing vision shortcuts.
CAG improves grounding to 96.7\% and task success by 36.7\%, showing more reliable goal targeting.

\textbf{OOD Generalization.}
In \emph{Cup/Cube/Basketball,} we evaluate zero-shot language grounding on OOD objects.
Although Cube and Basketball are unseen during finetuning, $\pi_{0.5}$ still achieves 50\% and 60\% success rates, showing strong zero-shot generalization with large-scale pretraining.
However, this is largely compromised by counterfactual failures: e.g., 4 out of 5 failures on Cube are counterfactual failures.
CAG mitigates this issue and further unlocks zero-shot generalization of VLAs.

\textbf{Summary.}
Although $\pi_{0.5}$ achieves high performance across all training tasks, it suffers from severe counterfactual failures.
CAG provides a simple yet effective solution across all evaluated dimensions, while preserving in-domain performance on original training tasks.
Additional real-world experiments on long-horizon reasoning, qualitative comparisons, and detailed per-task results are provided in Appendix~\ref{sec:real}.

\subsection{Analysis}

\begin{table}[t]
\centering
\begin{minipage}[t]{0.55\linewidth}
    \centering
    \scriptsize
    \setlength{\tabcolsep}{3pt}
    \renewcommand{\arraystretch}{1.05}
    \caption{\textbf{CAG with additional counterfactual data.}}
    \label{tab:cf_budget}
    \vspace{3pt}
    \begin{tabular}{c|cc|cc|cc|cc}
        \toprule
        \# Demos
        & \multicolumn{2}{c|}{Left}
        & \multicolumn{2}{c|}{Right}
        & \multicolumn{2}{c|}{Plate}
        & \multicolumn{2}{c}{Basket} \\
        & $\pi_{0.5}$ & +CAG
        & $\pi_{0.5}$ & +CAG
        & $\pi_{0.5}$ & +CAG
        & $\pi_{0.5}$ & +CAG \\
        \midrule
        0 & 0/10 & 0/10
          & 0/10 & 0/10
          & 3/10 & 8/10
          & 0/10 & 0/10 \\
        1 & 1/10 & 3/10
          & 3/10 & 4/10
          & 2/10 & 7/10
          & 2/10 & 6/10 \\
        2 & 4/10 & 8/10
          & 2/10 & 5/10
          & 8/10 & 9/10
          & 8/10 & 8/10 \\
        \bottomrule
    \end{tabular}
\end{minipage}
\hfill
\begin{minipage}[t]{0.44\linewidth}
    \centering
    \scriptsize
    \setlength{\tabcolsep}{4pt}
    \renewcommand{\arraystretch}{1.05}
    \caption{\textbf{Computational overhead.}}
    \label{tab:overhead}
    \vspace{8pt}
    \begin{tabular}{lcc}
        \toprule
        \textbf{Method} & \textbf{GPU Memory (GB)} & \textbf{Latency (ms)} \\
        \midrule
        $\pi_{0.5}$ & 8.5 & 80.0 $\pm$ 2.2 \\
        TF-CAG & 8.5 (1.0$\times$) & 126.6 $\pm$ 4.4 (1.6$\times$) \\
        VA-CAG & 16.5 (1.9$\times$) & 132.6 $\pm$ 1.9 (1.7$\times$) \\
        \bottomrule
    \end{tabular}
\end{minipage}
\vspace{-6pt}
\end{table}

\textbf{CAG Complements Counterfactual Training Data.}
To investigate whether CAG complements additional counterfactual training data, we finetune the base policy with 20 demonstrations for each $l \in L_o^{\text{in}}$ and 0, 1, or 2 demonstrations for each $l \in L_o^{\text{out}}$.
Tab.~\ref{tab:cf_budget} reports success over 10 trials on four counterfactual tasks (Left, Right, Plate, and Basket) with and without CAG.
The results show that CAG and additional counterfactual data provide complementary benefits: more counterfactual data generally improves performance, while CAG provides additional gains at the same data budget.

\textbf{Additional overhead of CAG.}
TF-CAG requires no additional training, while VA-CAG finetunes an additional vision-only branch (around 3 hours on LIBERO and 1 hour on real-world data using 8 H100 GPUs).
Tab.~\ref{tab:overhead} reports GPU memory usage and per-action-chunk inference latency during our real-world deployment.
Both variants run conditional and unconditional inference in parallel, retaining low latency (less than $2\times$ the baseline, around 130 ms per 15-step action chunk).

\section{Conclusion} 
\label{sec:conclusion}
We provide a systematic study of vision shortcuts and counterfactual failures in VLAs.
We introduce the first counterfactual benchmark LIBERO-CF, and propose Counterfactual Action Guidance (CAG), a simple yet effective dual-branch inference scheme that strengthens language conditioning in VLAs.
Extensive experiments in both simulation and the real world demonstrate the presence of vision shortcuts, and CAG effectively reduces counterfactual failures.

\textbf{Limitations.}
CAG is most effective when the base policy already has the necessary perception and low-level execution capability; if a skill is entirely absent, action-space guidance alone cannot create it. 
This scope is also a strength: it is orthogonal to model scaling, data augmentation, and architecture design.
Moreover, dual-branch inference introduces additional
computational overhead, while VA-CAG requires training
a separate vision-only prior.
Future work will extend to richer environments, longer-horizon tasks, and broader embodiments, and explore adaptive guidance based on uncertainty or visual-language disagreement.



\clearpage
\acknowledgments{We thank the anonymous reviewers and area chairs for their constructive feedback, which helped improve this work.}


\bibliography{example}  
\newpage
\section*{Appendix}
\appendix

The appendix includes three sections.
\begin{itemize}[leftmargin=*, noitemsep, topsep=0pt]
    \item Appendix~\ref{sec:supl_cag} further explains the practical significance of this paper.
    \item Appendix~\ref{sec:supl_sim} presents additional simulation experiments on an extra VLA (i.e., X-VLA~\cite{zheng2025xvla}), together with implementation details.
    \item Appendix~\ref{sec:real} includes extended real-world experiments on long-horizon reasoning, along with detailed scene descriptions and a complete breakdown of results across all evaluated scenes.
\end{itemize}

\section{Practical Significance}
\label{sec:supl_cag}
We study an underexplored OOD problem in real-life robotics: it is infeasible to anticipate every task a user may need, new objects and goals inevitably emerge after deployment, and a reliable VLA must handle them faithfully.
The core promise of VLAs is that large-scale pretraining should enable generalization beyond the finetuning distribution: a user finetunes VLA on a small set of tasks, expecting it to generalize beyond narrow scene-specific supervision.
This paper directly studies whether this promise holds: can a VLA faithfully follow counterfactual tasks (under-observed or OOD instructions, e.g., targeting an unseen object), or does it default to finetuned behaviors?
We believe this is a practically important gap that current in-distribution VLA studies overlook.

\section{Extended Simulation Experiments}
\label{sec:supl_sim}

\subsection{Additional Details in the Main Paper}

\textbf{VLA Baselines.} 
We discuss the language-following ability of representative and widely used VLAs. 
OpenVLA-OFT~\cite{kim2025openvla-oft} is a large-scale 7B VLA that uses Prismatic VLM~\cite{karamcheti2024prismatic} as the backbone, and is pretrained on the Open-X-Embodiment dataset.
$\pi_0$~\cite{black2024pi_0} is a 3B flow-matching-based VLA built upon PaliGemma~\cite{karamcheti2024prismatic}. 
$\pi_{0.5}$~\cite{intelligence2025pi_} is an improved variant of $\pi_0$ that adopts a hybrid pretraining procedure on large-scale robotic and non-robotic data, representing state-of-the-art performance.
For all models, we use their officially released weights for the LIBERO benchmark in our evaluations.

\textbf{Implementation Details of Sec.~\ref{sec:sim}.}
We conduct simulation experiments on the LIBERO-CF benchmark.
We run 50 trials for each task across all methods.
We use official finetuned VLA models on the LIBERO benchmark.
For VA models as $\pi_{\text{uncond}}$, we finetune on the LIBERO dataset using the default training configurations as its corresponding VLA, except that the language input is dropped.

\textbf{Variation in Counterfactual Failures across VLAs.}
In Tab.~\ref{tab:libero_cf}, we further observe that the prevalence of counterfactual failures and the effectiveness of CAG vary across different VLAs.
In particular, while all evaluated VLAs benefit from CAG, CAG reduces counterfactual failures on $\pi_{0.5}$ much more than on OpenVLA-OFT.
This suggests that $\pi_{0.5}$ exhibits a weaker form of vision shortcuts than OpenVLA-OFT (also in Tab.~\ref{tab:libero}), allowing CAG to more effectively amplify language-conditioned action shifts.
In contrast, OpenVLA-OFT demonstrates more rigid vision-dominated behaviors and requires stronger guidance to overcome its entrenched visual priors. 
Overall, these results indicate that counterfactual failures are not a binary phenomenon, but instead exist along a spectrum, influenced by model architecture, training strategy, and action modeling.

\begin{table*}[t]
    \caption{\textbf{Quantitative evaluation of X-VLA on LIBERO-CF.} Results are percentages (\%). F. denotes Faithful and B. denotes Biased.}
    \centering
    \small
    \setlength{\tabcolsep}{2pt}
    \renewcommand{\arraystretch}{1.05}
    \begin{tabular}{
      >{\centering\arraybackslash}p{1.8cm} |
      >{\centering\arraybackslash}p{1.4cm} |
      >{\centering\arraybackslash}p{0.8cm} |
      >{\centering\arraybackslash}p{0.8cm}
      >{\centering\arraybackslash}p{0.8cm}
      >{\centering\arraybackslash}p{0.8cm}
      >{\centering\arraybackslash}p{0.8cm}
      >{\centering\arraybackslash}p{0.8cm}
      >{\centering\arraybackslash}p{0.8cm}
      >{\centering\arraybackslash}p{0.8cm}
      >{\centering\arraybackslash}p{0.8cm}
      >{\centering\arraybackslash}p{0.8cm}
      >{\centering\arraybackslash}p{0.8cm}
    }
    \toprule
    \multirow{2}{*}{\textbf{Model}} &
    \multirow{2}{*}{\textbf{Metric}} &
    \multirow{2}{*}{\textbf{CAG}}
    & \multicolumn{2}{c}{\textbf{CF-Spatial}}
    & \multicolumn{2}{c}{\textbf{CF-Object}}
    & \multicolumn{2}{c}{\textbf{CF-Long}}
    & \multicolumn{2}{c}{\textbf{CF-OOD}}
    & \multicolumn{2}{c}{\textbf{Avg.}} \\
    \cmidrule(lr){4-5}\cmidrule(lr){6-7}\cmidrule(lr){8-9}
    \cmidrule(lr){10-11}\cmidrule(lr){12-13}
    & & & \textbf{F. $\uparrow$} & \textbf{B. $\downarrow$}
          & \textbf{F. $\uparrow$} & \textbf{B. $\downarrow$}
          & \textbf{F. $\uparrow$} & \textbf{B. $\downarrow$}
          & \textbf{F. $\uparrow$} & \textbf{B. $\downarrow$}
          & \textbf{F. $\uparrow$} & \textbf{B. $\downarrow$} \\
    \midrule

\multirow{6}{*}{X-VLA~\cite{zheng2025xvla}}
& \multirow{3}{*}{Grounding}
& /
& 22.1 & 72.1
& 1.4  & 70.4
& \textbf{98.2} & 23.3
& 27.3 & \textbf{58.4}
& 37.3 & 56.1 \\
& & TF
& 22.3 & 68.9
& 0.0  & 77.6
& 95.6 & \textbf{12.2}
& \textbf{40.9} & 63.7
& 39.7 & 55.6 \\
& & VA
& \textbf{24.4} & \textbf{68.8}
& \textbf{2.0}  & \textbf{59.6}
& 96.7 & 17.6
& 40.8 & 60.5
& \textbf{41.0} & \textbf{51.6} \\
\cmidrule(lr){2-13}
& \multirow{3}{*}{Success}
& /
& 3.2 & 58.9
& 0.0 & 50.2
& 47.6 & 9.1
& 4.4 & 38.8
& 13.8 & 39.3 \\
& & TF
& \textbf{8.9} & 62.3
& 0.0 & 56.6
& \textbf{59.1} & 3.8
& \textbf{10.1} & 47.9
& \textbf{19.5} & 42.6 \\
& & VA
& 8.7 & \textbf{53.7}
& 0.0 & \textbf{31.6}
& 51.1 & \textbf{0.9}
& 8.7 & \textbf{38.1}
& 17.1 & \textbf{31.1} \\
\bottomrule
\end{tabular}
\vspace{-12pt}
\label{tab:libero_cf_xvla}
\end{table*}

\subsection{Supplementary Results}
\textbf{Additional VLA Baseline.}
In addition to VLAs evaluated in the main paper, we include X-VLA~\cite{zheng2025xvla} as an additional state-of-the-art baseline.
X-VLA is a 0.9B VLA architecture that utilizes learnable soft-prompts to tackle cross-embodiment settings and learn across multiple domains ranging from simulation to real-world datasets.

\textbf{Results.}
In Tab.~\ref{tab:libero_cf_xvla}, we report additional quantitative results of X-VLA~\cite{zheng2025xvla} on the LIBERO-CF Benchmark.
X-VLA shows counterfactual failures similar to other VLAs, further confirming that this failure mode is not specific to particular architectures.
This reflects a strong reliance on scene-specific visual priors and limited robustness to alternative feasible instructions, consistent with the vision shortcut phenomenon observed in other VLAs.
Notably, both variants of CAG improve language grounding for X-VLA.
The training-free strategy (TF) increases average grounding from 37.3\% to 39.7\% and success from 13.8\% to 19.5\%, while reducing biased success on training tasks.
Using a Vision–Action prior (VA) increases average grounding to 41.0\% and success to 17.1\%.
Overall, these results further demonstrate that counterfactual failures are widespread across VLAs.
Importantly, CAG consistently mitigates such failures without modifying model architectures or pretrained checkpoints, highlighting its general applicability across diverse VLA designs.

\subsection{Implementation Details of VLA Baselines}
\textbf{OpenVLA-OFT}~\cite{kim2025openvla-oft}.
OpenVLA-OFT adopts a continuous action representation and employs an optimized finetuning recipe with parallel decoding and action chunking.
Unlike OpenVLA~\cite{kim24openvla}, actions are optimized via L1 regression instead of bin-wise classification.
We directly use the officially released LIBERO-finetuned checkpoint trained on the combined four task suites.
For the VA model, we provide empty language instructions to keep the remaining prompt structure unchanged, and follow the official LIBERO finetuning configuration.
We LoRA finetune OpenVLA-7B with a batch size of 64 for 300k steps using the Adam optimizer, a learning rate of $5\times10^{-4}$.
Since OpenVLA-OFT shows a stronger reliance on visual cues, we use a guidance scale of $\omega=3.0$ for both the TF and VA variants in our experiments.

$\pi_{0}$~\cite{black2024pi_0} \textbf{and} $\pi_{0.5}$~\cite{intelligence2025pi_}.
We directly adopt the officially released LIBERO-finetuned checkpoints for the VLA baselines.
For the VA model, we follow the same finetuning recipe but drop the language tokens.
We fully finetune the base models with a batch size of 32 for 30k steps using the Adam optimizer, a learning rate of $5\times10^{-5}$, and an EMA decay of 0.999.
For CAG, we use a guidance scale of $\omega=1.5$ for both TF and VA.

\textbf{X-VLA}~\cite{zheng2025xvla}.
For the VLA baseline, we finetune the officially released base model (X-VLA-Pt) on the LIBERO dataset in the absolute end-effector action space.
We use bf16 mixed precision, a batch size of 16, a learning rate of $1\mathrm{e}{-4}$ with a learning-rate coefficient of 0.1, and train for 50k iterations with 2k warmup steps, freezing the backbone for the first 1k steps.
For the VA model, we use the same finetuning configuration as the VLA baseline, but drop the language tokens so that the model is conditioned only on visual observations.
For CAG, we use a guidance scale of $\omega=1.5$ for both the TF and VA variants in all X-VLA experiments.

\section{Extended Real-world Experiments}
\label{sec:real}

\begin{figure}[t]
  \centering
  \includegraphics[width=0.5\linewidth]{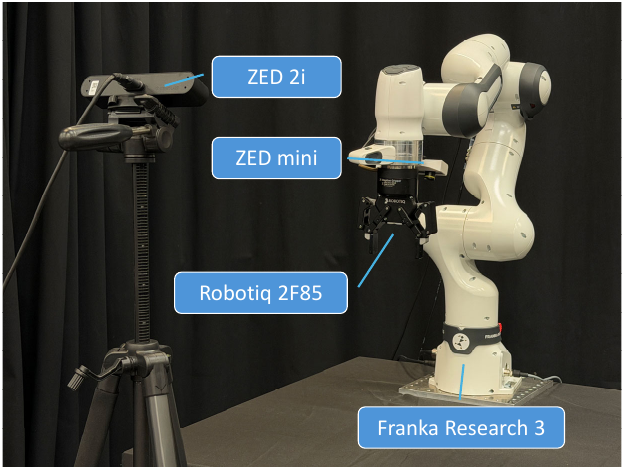}
  \caption{\textbf{Our real-world setup.} We conduct all real-world experiments using a Franka Research 3 robot arm with a Robotiq 2F-85 gripper, with a ZED~2i and a ZED~mini.}
  \label{fig:setup}
  \vspace{-8pt}
\end{figure}

\subsection{Details of Our Real-world Datasets}
We provide detailed task descriptions for each real-world scene in Tab.~\ref{tab:task}.
All expert demonstrations are collected via teleoperation using the DROID platform~\cite{khazatsky2024droid}.
For each in-domain task in $L_o^{\text{in}}$, we collect 20 expert demonstrations to ensure sufficient scene-specific supervision.
For each under-observed task in $L_o^{\text{out}}$, we only collect a single demonstration for minimal warm-up and treat these tasks as counterfactual during evaluation.
Our experiments focus on under-observed tasks to investigate counterfactual failures in VLAs and to validate the effectiveness of Counterfactual Action Guidance (CAG), while also reporting results on in-domain tasks to verify that CAG preserves performance on in-domain tasks.

\begin{figure*}[t]
  \centering
    \includegraphics[width=0.95\linewidth]{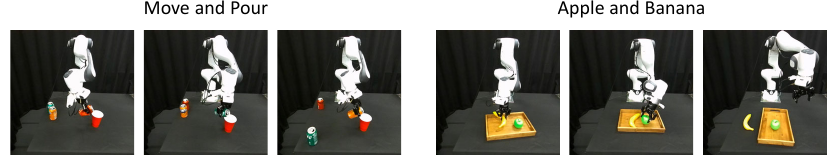}
    \par\vspace{5pt}
    \includegraphics[width=0.95\linewidth]{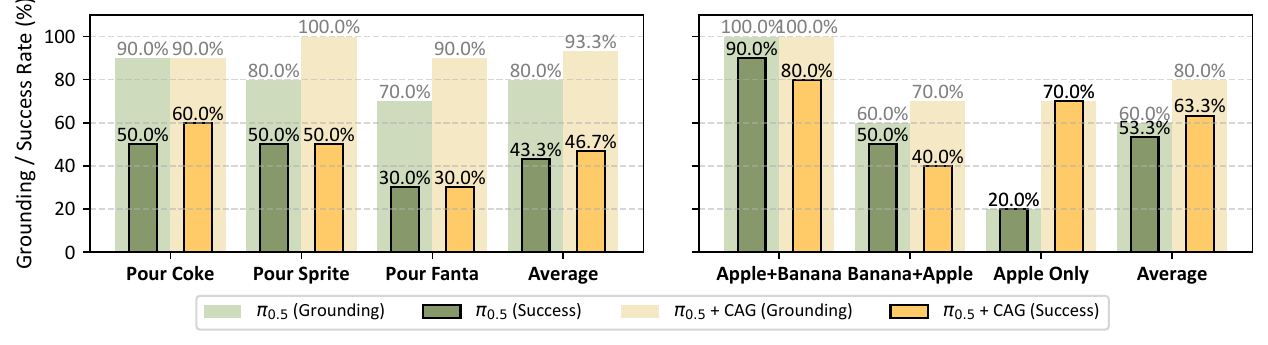}
    \caption{\textbf{Additional real-world experiments for long-horizon reasoning.} CAG improves the performance of $\pi_{0.5}$ across both scenes, consistent with the discussed dimensions in the main paper.}
    \label{fig:realworld_supl}
    \vspace{-16pt}
\end{figure*}

\begin{table}[t]
\centering
\scriptsize
\caption{\textbf{Real-world task descriptions across scenes.}
Each scene contains one in-domain instruction and two under-observed instructions.}
\label{tab:task}
\renewcommand{\arraystretch}{1.15}
\setlength{\tabcolsep}{3pt}

\begin{tabularx}{\textwidth}{
>{\centering\arraybackslash}p{2.2cm}
>{\centering\arraybackslash}p{2.8cm}
>{\centering\arraybackslash}p{3.2cm}
X
}
\toprule
\textbf{Dimension} & \textbf{Scene} & 
\textbf{In-domain Task} $L_o^{\text{in}}$ & 
\textbf{Under-observed Tasks} $L_o^{\text{out}}$ \\
\midrule

\multirow{2}{*}{Object Recognition}
& Coke / Sprite / Fanta
& Pick up the coke
& Pick up the sprite; pick up the fanta \\

& Tape / Mustard / Pringles
& Pick up the tape
& Pick up the mustard; pick up the pringles \\

\midrule

\multirow{2}{*}{Spatial Reasoning}
& Middle / Left / Right
& Pick up the cup in the middle
& Pick up the cup on the left; pick up the cup on the right \\

& Table / Plate / Bowl
& Pick up the corn can on the table
& Pick up the corn can on the plate; pick up the corn can in the bowl \\

\midrule

Goal Targeting
& Stack / Plate / Basket
& Stack the cups
& Put the cup on the plate; put the cup in the basket \\

\midrule

OOD Generalization
& Cup / Cube / Basketball
& Pick up the cup
& Pick up the Rubik's Cube; pick up the basketball \\

\midrule

\multirow{2}{*}{Long-Horizon Reasoning}
& Move and Pour
& Move the cup to the right and pour the coke into the cup
& Move the cup to the right and pour the fanta into the cup; move the cup to the right and pour the sprite into the cup \\

& Apple and Banana
& Put the apple on the tray and then put the banana on the tray
& Put the banana on the tray and then put the apple on the tray; put the apple on the tray \\

\bottomrule
\end{tabularx}
\vspace{-16pt}
\end{table}

\subsection{Additional Experiments for Long-Horizon Reasoning}
In the main paper, we study multiple aspects of language grounding in real-world settings, including: 1) \emph{Object Recognition}: Three distinct objects are present in the scene, and tasks differ by the target object specified in the instruction; 2) \emph{Spatial Reasoning}: Three identical objects are placed at different spatial locations, requiring the robot to distinguish based on spatial language; 3) \emph{Goal Targeting}: Three tasks are defined for the same object, requiring the understanding of different goal specifications; and 4) \emph{Out-of-Distribution (OOD) Generalization}: Distinguish one familiar and two entirely unseen objects in the finetuning dataset.
\textbf{In this appendix, we further extend our evaluation to an additional dimension: 5) \emph{Long-Horizon Reasoning}.}
This examines the ability of VLAs to follow multi-step instructions and maintain language grounding over extended action horizons.

\textbf{Task Design.}
As in other dimensions, each scene includes three possible tasks, one belongs to well-learned in-domain tasks $L_o^{\text{in}}$ with sufficient expert demonstrations, where the remaining two correspond to under-observed tasks $L_o^{\text{out}}$ with counterfactual instructions.
We design two scenes: \emph{Move and Pour} and \emph{Apple and Banana} for long-horizon reasoning. We present our results in Fig.~\ref{fig:realworld_supl}.

\textbf{\emph{Move and Pour}.}
We study whether VLAs can correctly ground the instructed objects during long-horizon, multi-step instructions.
The in-domain task is “Move the cup to the right and pour the coke into the cup”, while counterfactual tasks replace the poured object with Sprite or Fanta.
In Fig.~\ref{fig:realworld_supl}, $\pi_{0.5}$ shows clear counterfactual failures: It frequently pours the wrong drink, typically defaulting to the training-task object (coke).
This indicates that counterfactual failures persist even when object grounding is required at later stages of long-horizon execution, and that VLAs tend to over-rely on memorized training-task objects when resolving ambiguity across steps.
Notably, these failures occur more frequently than in the corresponding short-horizon object recognition task (\emph{Coke/Sprite/Fanta}), suggesting that vision shortcuts compound over time.
CAG effectively strengthens language conditioning throughout the execution horizon, preventing early visual priors from dominating later decision points.

\textbf{\emph{Apple and Banana}.}
We investigate whether VLAs can correctly understand both \emph{order} and \emph{cardinality} of multi-step instructions.
More specifically, while the in-domain task is ``Put the apple on the tray and then put the banana on the tray", we design two counterfactual tasks: 1) reversing the order (``Put the banana on the tray and then put the apple on the tray"), and 2) a partial instruction that requires only the first step (``Put the apple on the tray").
In Fig.~\ref{fig:realworld_supl}, $\pi_{0.5}$ shows clear counterfactual failures in both settings.
For the reversed-order instruction, it often follows the original training order, placing the apple first despite explicit language specifying the opposite order.
For the partial instruction, it frequently over-executes by completing the full training sequence, placing both objects instead of stopping after the apple.
These failures indicate that counterfactual failures in VLAs can manifest as \emph{temporal and compositional errors}, where the model tends to fall back on memorized action sequences rather than faithfully following the instructions.
We observe that CAG effectively mitigates these failures.
Both grounding and success rates improve across all long-horizon variants, demonstrating that language guidance more effectively reweights the action distribution at each step.
In particular, CAG significantly improves success on the ``apple-only'' task, showing that it can suppress false continuation of learned action sequence and enforce instruction termination.
Overall, these results suggest that counterfactual failures become more severe over long horizons, and that CAG helps maintain language grounding throughout multi-step execution.

\textbf{Summary.}
In Fig.~\ref{fig:real-world-object-spatial-goal} and Fig.~\ref{fig:realworld_supl}, although $\pi_{0.5}$ achieves high performance across all training tasks, it suffers from severe counterfactual failures.
CAG provides a simple yet effective solution across all evaluated dimensions, while preserving in-domain performance on original training tasks.

\subsection{Detailed Results}
\begin{figure*}[t]
  \centering
    \includegraphics[width=0.9\linewidth]{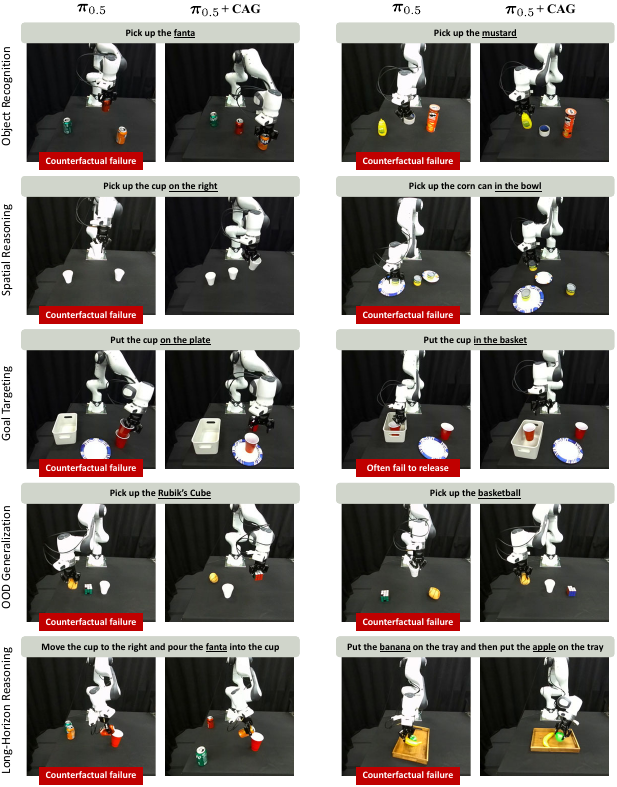}
    \caption{\textbf{Qualitative comparisons with and without Counterfactual Action Guidance.} }
    \vspace{-8pt}
    \label{fig:realworld_comparison}
\end{figure*}

\begin{figure*}[t]
  \centering
    \includegraphics[width=1.0\linewidth]{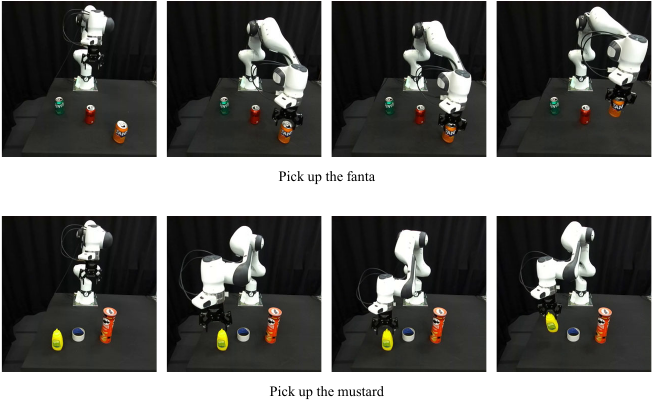}
    \caption{\textbf{Real-world rollouts using Counterfactual Action Guidance for object recognition.}}
    \label{fig:realworld_object}
\end{figure*}

\begin{figure*}[t]
  \centering
    \includegraphics[width=1.0\linewidth]{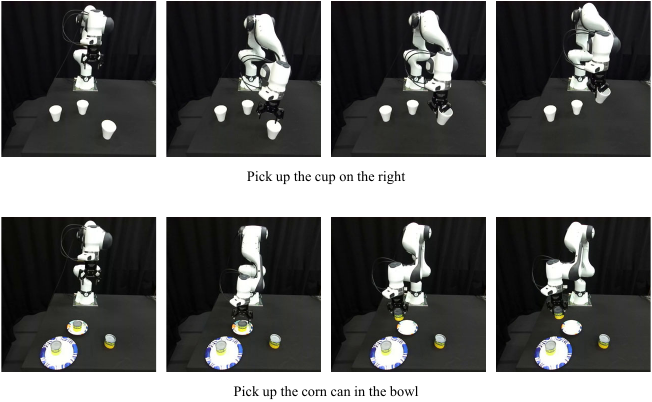}
    \caption{\textbf{Real-world rollouts using Counterfactual Action Guidance for spatial reasoning.} }
    \label{fig:realworld_spatial}
\end{figure*}

\begin{figure*}[t]
  \centering
    \includegraphics[width=1.0\linewidth]{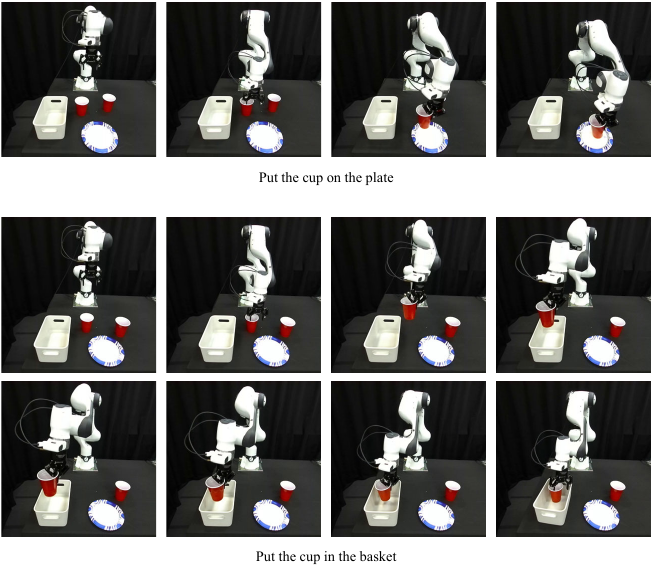}
    \caption{\textbf{Real-world rollouts using Counterfactual Action Guidance for goal targeting.} }
    \label{fig:realworld_goal}
\end{figure*}

\begin{figure*}[t]
  \centering
    \includegraphics[width=1.0\linewidth]{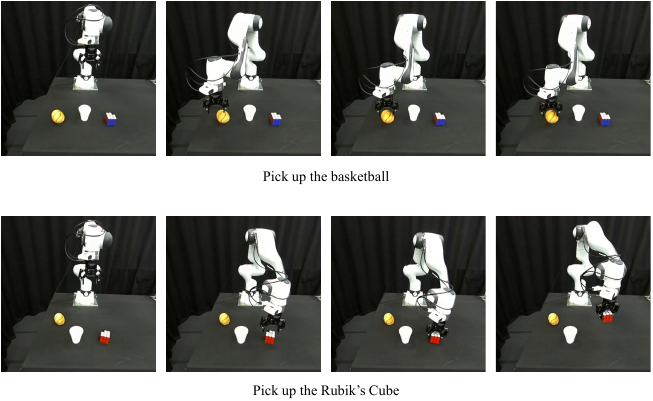}
    \caption{\textbf{Real-world rollouts using Counterfactual Action Guidance for out-of-distribution generalization.} }
    \label{fig:realworld_ood}
\end{figure*}

\begin{figure*}[t]
  \centering
    \includegraphics[width=1.0\linewidth]{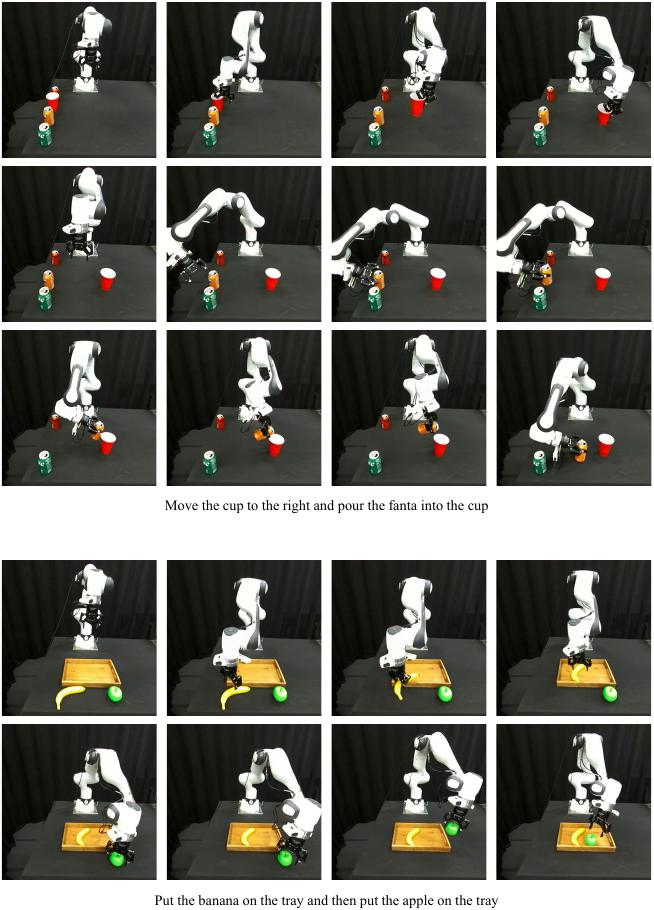}
    \caption{\textbf{Real-world rollouts using Counterfactual Action Guidance for long-horizon reasoning.} }
    \label{fig:realworld_long}
\end{figure*}

\textbf{Qualitative results.}
In Fig.~\ref{fig:realworld_comparison}, we present the comparisons with and without Counterfactual Action Guidance.
We further provide qualitative results from real-world rollouts using Counterfactual Action Guidance, including object recognition in Fig.~\ref{fig:realworld_object}, spatial reasoning in Fig.~\ref{fig:realworld_spatial}, goal targeting in Fig.~\ref{fig:realworld_goal}, out-of-distribution generalization in Fig.~\ref{fig:realworld_ood}, and long-horizon reasoning in Fig.~\ref{fig:realworld_long}.

\textbf{Quantitative results.} 
To further analyze the failure patterns, we report the full breakdown of our real-world evaluation results in Tab.~\ref{tab:realworld_part1} and Tab.~\ref{tab:realworld_part2}.
For each task instruction, we conduct 10 trials and report each entry as successful executions / total attempts.
For example, in the \emph{Coke/Sprite/Fanta} scene with $\pi_{0.5}$, when given the instruction ``Pick up the fanta", the robot attempts the correct task 9 times and succeeds 6 times (reported as ``6/9"), corresponding to a 90\% grounding rate and a 60\% task success rate. 
Notably, there is also one trial in which the robot instead grasps the coke and succeeds (reported as ``1/1"), illustrating a concrete instance of a counterfactual failure.
The “Fail” columns record trials in which the robot remains idle or does not show any clear grounding toward any specific object.
This detailed breakdown enables fine-grained analysis of failure modes, allowing us to explicitly distinguish correct instruction following behavior from biased execution toward well-learned training tasks.

\begin{table*}[t]
\centering
\caption{\textbf{Breakdown results in real-world experiments (Part 1).}
Each entry is reported as successful executions / total attempts.
Rows denote the instructed target; columns denote the executed outcome.}
\renewcommand{\arraystretch}{1.15}
\small

\begin{tabular}{C{1.5cm} C{2.0cm} C{1.1cm} C{1.1cm} C{1.1cm} C{0.9cm}}
\toprule

\multicolumn{6}{c}{\textbf{Coke / Sprite / Fanta}} \\
\cmidrule(lr){1-6}
\textbf{Model} & \textbf{Instruction} & \textbf{Coke} & \textbf{Sprite} & \textbf{Fanta} & \textbf{Fail} \\
\cmidrule(lr){1-6}

\multirow{3}{*}{$\pi_{0.5}$}
 & ``Coke''   & 8/10 &      &      &      \\
 & ``Sprite'' &      & 6/10 &      &      \\
 & ``Fanta''  & 1/1  &      & 6/9  &      \\

\cmidrule(lr){1-6}
\multirow{3}{*}{$\pi_{0.5}$ + CAG}
 & ``Coke''   & 9/10 &      &      &      \\
 & ``Sprite'' &      & 8/10 &      &      \\
 & ``Fanta''  &      &      & 7/10 &      \\

\midrule

\multicolumn{6}{c}{\textbf{Tape / Mustard / Pringles}} \\
\cmidrule(lr){1-6}
\textbf{Model} & \textbf{Instruction} & \textbf{Tape} & \textbf{Mustard} & \textbf{Pringles} & \textbf{Fail} \\
\cmidrule(lr){1-6}

\multirow{3}{*}{$\pi_{0.5}$}
 & ``Tape''     & 10/10 &      &      &      \\
 & ``Mustard''  & 3/3   & 3/6  & 0/1  &      \\
 & ``Pringles'' &       & 1/2  & 2/8  &      \\

\cmidrule(lr){1-6}
\multirow{3}{*}{$\pi_{0.5}$ + CAG}
 & ``Tape''     & 9/10 &      &      &      \\
 & ``Mustard''  &      & 6/10 &      &      \\
 & ``Pringles'' &      &      & 4/10 &      \\

\midrule

\multicolumn{6}{c}{\textbf{Middle / Left / Right}} \\
\cmidrule(lr){1-6}
\textbf{Model} & \textbf{Instruction} & \textbf{Middle} & \textbf{Left} & \textbf{Right} & \textbf{Fail} \\
\cmidrule(lr){1-6}

\multirow{3}{*}{$\pi_{0.5}$}
 & ``Middle'' & 9/10 &      &      &      \\
 & ``Left''   & 8/8  & 1/2  &      &      \\
 & ``Right''  & 7/7  &      & 3/3  &      \\

\cmidrule(lr){1-6}
\multirow{3}{*}{$\pi_{0.5}$ + CAG}
 & ``Middle'' & 10/10 &      &      &      \\
 & ``Left''   & 6/7   & 3/3  &      &      \\
 & ``Right''  & 4/5   & 1/1  & 4/4  &      \\

\midrule

\multicolumn{6}{c}{\textbf{Table / Plate / Bowl}} \\
\cmidrule(lr){1-6}
\textbf{Model} & \textbf{Instruction} & \textbf{Table} & \textbf{Plate} & \textbf{Bowl} & \textbf{Fail} \\
\cmidrule(lr){1-6}

\multirow{3}{*}{$\pi_{0.5}$}
 & ``Table'' & 7/9 & 1/1 &     &      \\
 & ``Plate'' &     & 6/8 & 2/2 &      \\
 & ``Bowl''  &     & 4/5 & 4/5 &      \\

\cmidrule(lr){1-6}
\multirow{3}{*}{$\pi_{0.5}$ + CAG}
 & ``Table'' & 8/10 &      &      &      \\
 & ``Plate'' &      & 7/7  & 3/3  &      \\
 & ``Bowl''  &      & 3/4  & 6/6  &      \\

\bottomrule
\end{tabular}
\label{tab:realworld_part1}
\end{table*}

\begin{table*}[t]
\centering
\caption{\textbf{Breakdown results in real-world experiments (Part 2).}
Each entry is reported as successful executions / total attempts.
Rows denote the instructed target; columns denote the executed outcome.}
\renewcommand{\arraystretch}{1.15}
\small

\begin{tabular}{C{1.5cm} C{2.2cm} C{1.1cm} C{1.1cm} C{1.3cm} C{0.9cm}}
\toprule

\multicolumn{6}{c}{\textbf{Stack / Plate / Basket}} \\
\cmidrule(lr){1-6}
\textbf{Model} & \textbf{Instruction} & \textbf{Stack} & \textbf{Plate} & \textbf{Basket} & \textbf{Fail} \\
\cmidrule(lr){1-6}

\multirow{3}{*}{$\pi_{0.5}$}
 & ``Stack''  & 5/9 & 1/1 &      &      \\
 & ``Plate''  & 0/1 & 2/7 &      & 2    \\
 & ``Basket'' &     &     & 2/10 &      \\

\cmidrule(lr){1-6}
\multirow{3}{*}{$\pi_{0.5}$ + CAG}
 & ``Stack''  & 7/10 &      &      &      \\
 & ``Plate''  &      & 7/10 &      &      \\
 & ``Basket'' &      &      & 6/9  & 1    \\

\midrule

\multicolumn{6}{c}{\textbf{Cup / Cube / Basketball}} \\
\cmidrule(lr){1-6}
\textbf{Model} & \textbf{Instruction} & \textbf{Cup} & \textbf{Cube} & \textbf{Basketball} & \textbf{Fail} \\
\cmidrule(lr){1-6}

\multirow{3}{*}{$\pi_{0.5}$}
 & ``Cup''        & 9/10 &     &     &      \\
 & ``Cube''       & 0/1  & 5/6 & 3/3 &      \\
 & ``Basketball'' & 3/3  &     & 6/7 &      \\

\cmidrule(lr){1-6}
\multirow{3}{*}{$\pi_{0.5}$ + CAG}
 & ``Cup''        & 9/10 &     &     &      \\
 & ``Cube''       &      & 8/9 & 1/1 &      \\
 & ``Basketball'' & 1/2  &     & 7/8 &      \\

\midrule

\multicolumn{6}{c}{\textbf{Move and Pour}} \\
\cmidrule(lr){1-6}
\textbf{Model} & \textbf{Instruction} & \textbf{Coke} & \textbf{Sprite} & \textbf{Fanta} & \textbf{Fail} \\
\cmidrule(lr){1-6}

\multirow{3}{*}{$\pi_{0.5}$}
 & ``Coke''   & 5/9 & 0/1 &     &      \\
 & ``Sprite'' & 1/2 & 5/8 &     &      \\
 & ``Fanta''  & 2/3 &     & 3/7 &      \\

\cmidrule(lr){1-6}
\multirow{3}{*}{$\pi_{0.5}$ + CAG}
 & ``Coke''   & 6/9 &     & 1/1 &      \\
 & ``Sprite'' &     & 5/10&     &      \\
 & ``Fanta''  &     & 1/1 & 3/9 &      \\

\midrule

\multicolumn{6}{c}{\textbf{Apple and Banana}} \\
\cmidrule(lr){1-6}
\textbf{Model} & \textbf{Instruction} & \textbf{A+B} & \textbf{B+A} & \textbf{A} & \textbf{Fail} \\
\cmidrule(lr){1-6}

\multirow{3}{*}{$\pi_{0.5}$}
 & ``Apple+Banana'' & 9/10 &     &     &      \\
 & ``Banana+Apple'' & 4/4  & 5/6 &     &      \\
 & ``Apple''      & 8/8  &     & 2/2 &      \\

\cmidrule(lr){1-6}
\multirow{3}{*}{$\pi_{0.5}$ + CAG}
 & ``Apple+Banana'' & 8/10 &     &     &      \\
 & ``Banana+Apple'' & 2/3  & 4/7 &     &      \\
 & ``Apple''        & 3/3  &     & 7/7 &      \\

\bottomrule
\end{tabular}
\label{tab:realworld_part2}
\end{table*}
\end{document}